\documentclass[runningheads]{llncs}

\usepackage{eccv}

\usepackage{eccvabbrv}

\usepackage{graphicx}
\usepackage{booktabs}
\usepackage{caption}
\usepackage{wrapfig}
\usepackage{lscape} 
\usepackage{enumitem}
\usepackage{ulem}

\usepackage[accsupp]{axessibility}  

\usepackage[pagebackref,breaklinks,colorlinks]{hyperref}
\usepackage{hyperref}

\usepackage{orcidlink}
\usepackage[whole]{bxcjkjatype}

\usepackage{bm}
\usepackage{amssymb}
\usepackage{amsmath}
\usepackage{multirow}
\usepackage{pifont}
\newcommand{\xmark}{\ding{55}}%
\usepackage{colortbl}
\usepackage{xcolor}
\usepackage{comment}

\usepackage{orcidlink}

\newif\ifcolortext
\colortextfalse 

\ifcolortext
\newcommand{\ryu}[1]{{\color{purple}{#1}}}
\newcommand{\cRyo}[1]{{\color{red}{#1}}}
\newcommand{\hk}[1]{{\color{blue}{#1}}}
\newcommand{\chrup}[1]{{\color{red}{#1}}}
\newcommand{\na}[1]{{\color{magenta}{#1}}}
\newcommand{\ya}[1]{{\color{orange}{#1}}}
\newcommand{\iro}[1]{{\color{cyan}{#1}}}
\newcommand{\rn}[1]{{\color{green}{#1}}}

\newcommand{\SM}[1]{{\color{magenta}{#1}}} 
\else
\newcommand{\ryu}[1]{{\color{black}{#1}}}

\newcommand{\cRyo}[1]{{\color{black}{#1}}}
\newcommand{\hk}[1]{{\color{black}{#1}}}
\newcommand{\chrup}[1]{{\color{black}{#1}}}
\newcommand{\ya}[1]{{\color{black}{#1}}}
\newcommand{\iro}[1]{{\color{black}{#1}}}
\newcommand{\rn}[1]{{\color{black}{#1}}}
\newcommand{\SM}[1]{{\color{black}{#1}}} 
\newcommand{\na}[1]{{\color{black}{#1}}}
\fi

\usepackage{xr}
\makeatletter
\newcommand*{\addFileDependency}[1]{
 \typeout{(#1)}
 \@addtofilelist{#1}
 \IfFileExists{#1}{}{\typeout{No file #1.}}
}
\makeatother
\newcommand*{\myexternaldocument}[1]{
 \externaldocument{#1}
 \addFileDependency{#1.tex}
 \addFileDependency{#1.aux}
}
\myexternaldocument{supp_EN}

\makeatletter
\def\blfootnote{\gdef\@thefnmark{}\@footnotetext}
\makeatother

\begin{document}

\title{
\chrup{Scaling Backwards:\\Minimal Synthetic Pre-training?}
}

\titlerunning{Scaling Backwards}

\author{Ryo Nakamura\inst{1,*}
\orcidlink{0009-0009-0858-1229} 
\and
Ryu Tadokoro\inst{2,*}
\orcidlink{0009-0001-9473-3832}
\and
Ryosuke Yamada\inst{1}
\orcidlink{0000-0002-2154-8230}
\and\\
Yuki M. Asano \inst{3}\orcidlink{0000-0002-8533-4020}
\and
Iro Laina\inst{4}\orcidlink{0000-0001-8857-7709}
\and
Christian Rupprecht\inst{4}\orcidlink{0000-0003-3994-8045}
\and
Nakamasa Inoue\inst{5}\orcidlink{0000-0002-9761-4142}
\and
Rio Yokota\inst{5}\orcidlink{0000-0001-7573-7873}
\and
Hirokatsu Kataoka\inst{1}\orcidlink{0000-0001-8844-165X}
}

\authorrunning{R. Nakamura and R. Tadokoro et al.}

\institute{National Institute of Advanced Industrial Science and Technology (AIST)
\and Tohoku University \and
University of Amsterdam
\and
University of Oxford
\and
Tokyo Institute of Technology\\
}

\maketitle

\vspace{-15pt}

\begin{abstract}
Pre-training and transfer learning are an important building block of current computer vision systems. While pre-training is usually performed on large real-world image datasets, in this paper we ask whether this is truly necessary. To this end, we search for a minimal, purely synthetic pre-training dataset that allows us to achieve performance similar to the 1 million images of ImageNet-1k. We construct such a dataset from a single fractal with perturbations. With this, we contribute three main findings. (i) We show that pre-training is effective even with minimal synthetic images, with performance on par with large-scale pre-training datasets like ImageNet-1k for full fine-tuning. (ii) We investigate the single parameter with which we construct artificial categories for our dataset. We find that while the shape differences can be indistinguishable to humans, they are crucial for obtaining strong performances. (iii) Finally, we investigate the minimal requirements for successful pre-training. Surprisingly, we find that a substantial reduction of synthetic images from 1k to 1 can even lead to an \textit{increase} in pre-training performance, a motivation to further investigate ``scaling backwards''. Finally, we extend our method from synthetic images to real images to see if a single real image can show similar pre-training effect through shape augmentation. We find that the use of grayscale images and affine transformations allows even real images to ``scale backwards''. The code is available at \url{https://github.com/SUPER-TADORY/1p-frac}.

\end{abstract}

\section{Introduction}
\label{sec:intro}

\blfootnote{$^*$ These authors contributed equally}

\vspace{-7pt}
\hk{In image recognition,} pre-training allows discovering fundamental visual representations for downstream task applications. Pre-training enhances the performance of visual tasks and enables the use of small-scale task-specific datasets. \hk{Recently,} pre-training has been used as a key technology to construct foundation models trained on \hk{massive datasets with over hundreds of millions of images.} 
In some cases, a foundation model enables \ryu{adaptation of} zero-shot recognition without the need for additional data. 

Pre-training is often interpreted as discovering universal structures in large-scale datasets that later facilitate adaptation to down-stream tasks.
In this paper, we challenge this interpretation by providing a \textit{minimal} pre-training dataset, generated from a single fractal, that achieves similar downstream performance. 
\ya{At the heart of this investigation is the question of whether pre-training might simply be a better weight initialization rather than the discovery of useful visual concepts.
If true, performing expensive pre-training with hundreds of millions of images might not be necessary. 
}
\iro{This additionally frees pre-training from licensing or ethical issues.}

Since the rise of deep neural networks, the ImageNet dataset~\cite{DengCVPR2009_ImageNet} has been one of the most commonly used pre-training datasets. Originally, \hk{pre-training} has been conducted through supervised learning (SL) with human-provided teacher labels. However, it has become clear that \hk{pre-training} can also be achieved without human-provided labels, through self-supervised learning (SSL)~\cite{DoerschICCV2015,ZhangECCV2016,NorooziECCV2016,NorooziCVPR2018,GidarisICLR2018,HeCVPR2020,chen2020mocov2,ChenICCV2021_mocov3,ChenICML2020,ChenCVPR2021_simsiam,CaronICCV2021_dino,HeCVPR2022}. 

\begin{figure*}[t]
 \centering
\includegraphics[width=\linewidth]
{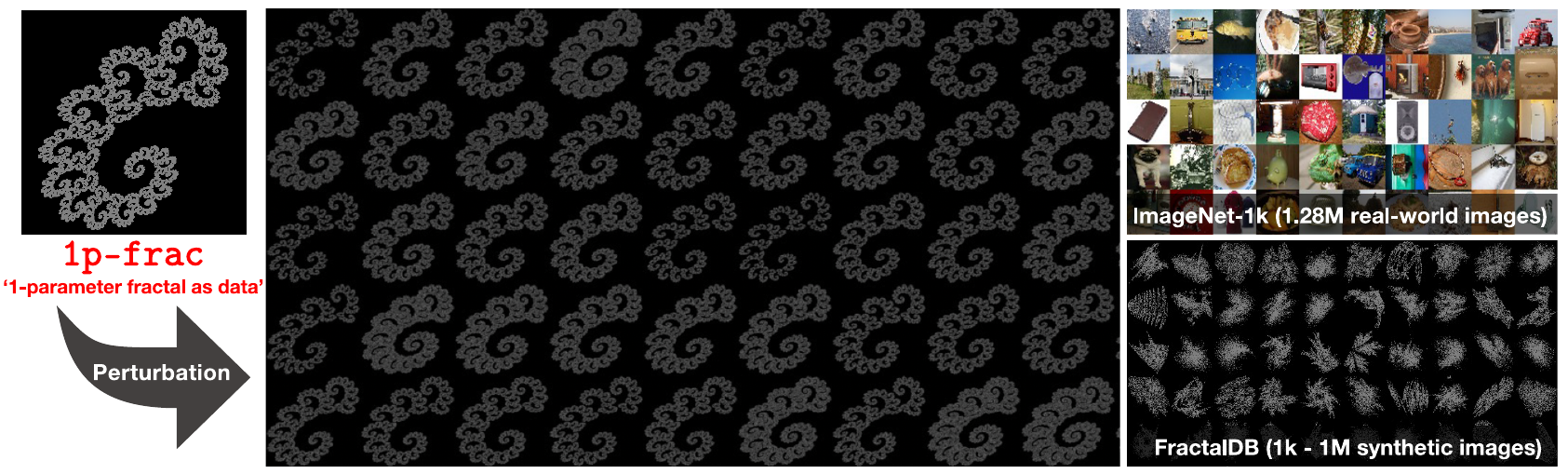}
\caption{
\na{
Comparison of ImageNet-1k, FractalDB and \texttt{1p-frac} (ours).
\texttt{1p-frac} consists of only a single fractal for pre-training.
With \texttt{1p-frac}, neural networks learn to classify perturbations applied to the fractal.
\rn{In our study  “single” means a very narrow distribution over parameters that leads to images that are roughly equivalent from a human visual perspective.}
While the shape differences of perturbed images can be indistinguishable to humans, models pre-trained on \texttt{1p-frac} achieve comparable performance with those pre-trained on ImageNet-1k or FractalDB.
}}
 \label{fig:teaser}
\end{figure*}

In this context, Asano \textit{et al.}~\cite{asano2020a} successfully acquired visual representations while dramatically reducing the number of required images.
They concluded that SSL can yield sufficient image representations even with a single training example, \ya{but only for} earlier layers of a recognition model.
However, it is unclear how these findings translate to modern architectures and representation learning methods. \hk{Based on this, vision transformers (ViT)~\cite{DosovitskiyICLR2021} have been pre-trained with only $2{,}040$ real images~\cite{ViT2040} through an instance discrimination learning signal.}

More recently, \ya{it has been shown} that basic visual representations can be acquired without using real images and human-provided labels. The trend of artificially generated labeled images is on the rise for synthetic pre-training~\cite{TobinIROS2017,BaradadNeurIPS2021,KataokaIJCV2022,KataokaCVPR2022}. Formula-driven supervised learning (FDSL) generates images from generative \hk{formulas}, and labels from their parameters~\cite{KataokaIJCV2022}. Under the FDSL framework, one can adapt the \hk{synthetic} pre-training image dataset by altering formulas~\cite{InoueICPR2020,KataokaICCV2021WS,KataokaWACV2022,TadokoroBMVC2023}. 
\iro{While a million-order image dataset was constructed in FractalDB~\cite{KataokaIJCV2022}, our findings suggest that synthetic pre-training can be reduced to significantly fewer} fractal images.

\ya{Motivated by these findings}, we believe it is possible to \hk{find} the key essence of \hk{pre-training for image recognition}. 
ViT training can be done with as few as $1{,}000$ artificially generated images~\cite{NakamuraICCV2023}. Here, we believe that equivalent performance can be achieved with even fewer images. 
This consideration is undoubtedly important as we approach a minimal \hk{synthetic pre-training} dataset in \hk{image recognition}, which \hk{goes against} the trend of foundation models toward increasing the dataset scale.

In \ryu{the present} paper, we thus introduce a \hk{\textit{minimal} synthetic} dataset,
namely 1-parameter Fractal as Data (\texttt{1p-frac}), \na{which consists of a single fractal as shown in Figure~\ref{fig:teaser}, as well as a loss function for pre-training with it}.
Our \na{contributions} regarding \hk{minimal synthetic} pre-training are as follows:

\noindent\textbf{Ordinal minimalism:} \na{We introduce the locally perturbed cross entropy (LPCE) loss for pre-training with a single fractal. It utilizes perturbed fractal images for training, where neural networks learn to classify small perturbations.}
In experiments, we demonstrate that pre-training can be performed even with a single fractal. The pre-training effect of \texttt{1p-frac} is \na{comparable} to \ryu{that of a} million-scale labeled image dataset.

\noindent\textbf{Distributional minimalism:}
We introduce the locally integrated empirical (LIEP) distribution $p_{\Delta}$ that has a controllable perturbation degree $\Delta$ to investigate the minimal support of the probability density distribution of synthesized images.
We observed positive pre-training effects even with a small $\Delta$ producing shape differences that cannot be distinguished by humans.
We also show that if $\Delta$ is too small
the visual pre-training collapses. From these observations, we \ya{establish the general bounds}
for generating good pre-training images from a mathematical formula.

\noindent\textbf{Instance minimalism:} 
\ryu{Based on the experimental results, the synthetic images should not simply contain complex shapes. The use of recursive image patterns, similar to objects in nature, should be applied in visual pre-training.}
\hk{Experiments with augmented categories from a real image have shown that good pre-training effects can be achieved by performing ``affine transformations on edge-emphasized objects in grayscale''. These operations are found to be almost synonymous with the configuration of the proposed \texttt{1p-frac}.}

\begin{wraptable}[6]{r}{50mm}
   \centering
   \setlength{\tabcolsep}{4pt}
   \vspace{-1.3cm} 
   \caption{\hk{Scaling backwards in synthetic pre-training (Accuracies on CIFAR-100, Real: ImageNet, Synth: Fractal images).}}
    \begin{tabular}{l|ccc}
        \toprule[0.8pt]
        Type$\backslash$\#Img & 1 & 1k & 1M \\
        \midrule[0.5pt]
        Real & N/A & 76.9 & 85.5  \\
        \rowcolor[gray]{0.8} Synth & 84.2 & 84.0 & 81.6 \\
        \bottomrule
    \end{tabular}
    \label{tab:scaling_backwards}
\end{wraptable}

In summary, \iro{we significantly reduce the size of the pre-training dataset from 1M images (fractal database (FractalDB)~\cite{KataokaIJCV2022}) or 1k images (one-instance fractal database (OFDB)~\cite{NakamuraICCV2023}) down to 1 and show that this even improves the pre-training effect (Table~\ref{tab:scaling_backwards}), which motivates ``scaling backwards''.}  


\section{Related Work}

\ryu{The present study focuses primarily} on finding \ryu{the} minimal pre-training dataset with artificially generated images. 

\paragraph{\hk{Pre-training in image recognition.}} Pre-training in image recognition has demonstrated significant success with vast amounts of labeled images. Pre-training helps neural networks to acquire fundamental visual representations which are essential for improving \ryu{the} performance \ryu{of} diverse downstream tasks in image recognition. \ryu{In particular}, researchers have \ryu{used} larger-scale image datasets for \hk{supervised learning (SL)}, starting from million-scale datasets \hk{(e.g., ImageNet~\cite{DengCVPR2009_ImageNet}, Places~\cite{ZhouTPAMI2017_Places})}. \ryu{More recently}, 
the largest datasets used are  
\iro{reaching billion-scale} \hk{(e.g., IG-3.5B~\cite{MahajanECCV2018_ig3.5b}, JFT-300M/3B/4B~\cite{SunICCV2017_jft300m,ZhaiCVPR2022_ScalingViT,DehghaniICML2023_ViT-22B})}. 

Pre-training can be also achieved without any human intervention.
One of the most effective approaches is self-supervised learning (SSL~\cite{DoerschICCV2015,ZhangECCV2016,NorooziECCV2016,NorooziCVPR2018,GidarisICLR2018,HeCVPR2020,chen2020mocov2,ChenICCV2021_mocov3,ChenICML2020,ChenCVPR2021_simsiam,CaronICCV2021_dino,HeCVPR2022}), which has been developed in order to alleviate the manual annotations on large-amount of images. SSL has been studied as an expected alternative to \hk{supervised} pre-training \iro{and has sometimes been shown to surpass supervised pre-training}. 

Despite the development of \hk{supervised and self-supervised pre-training}, the number of real images used in visual pre-training continues to increase.
\iro{Exploratory research questions such as \hk{``what is a minimal synthetic pre-training dataset?''} have received less attention.}
However, they could give us more insights on what is learned during pre-training and how it could be improved.

\paragraph{Pre-training with limited data.} \ryu{In order to} address ethical issues such as privacy and fairness associated with real images, \hk{a pre-training with a very limited number of real images and} synthetic pre-training without any real images \hk{have} been on the rise. In this context, synthetic pre-training with copy-and-paste learning~\cite{DwibediICCV2017,RemezECCV2018}, domain randomization from 3D graphics~\cite{TobinIROS2017,SundermeyerECCV2018}, learning from primitive patterns~\cite{BaradadNeurIPS2021,BaradadNeurIPS2022,SharmaCVPR2024}, and dataset distillation~\cite{WangarXiv2018_datasetdistillation,WangCVPR2022_datasetdistillation,ZhaoICLR2021_datasetcondensation} have been recognized as promising approaches to \ryu{acquiring} visual representations. More recently, a sophisticated approach, formula-driven supervised learning (FDSL)~\cite{KataokaIJCV2022,KataokaCVPR2022}, enabled \ryu{automatic} construct\ryu{ion of a} labeled image dataset without any real images \ryu{or} human annotations. FDSL can generate images and their corresponding labels \ryu{using} a mathematical formula. Along \ryu{these lines}, One-instance FractalDB (OFDB) has successfully pre-trained a \hk{ViT} using only $1{,}000$ synthetic images generated from a \hk{mathematical formula}, without \ryu{using} real images~\cite{NakamuraICCV2023}. Moreover, several studies have been reported in the \ryu{context} of extremely limited data~\cite{asano2020a,ViT2040,venkataramanan2023imagenet}. \ryu{In particular}, Asano \textit{et al.} have shown that learning visual representations is possible using a single image for self-supervised learning. Venkataramanan \textit{et al.} achieved a comparable method to ImageNet pre-training with only 10 walking videos and their cropped images. 

\ryu{By analyzing the acquisition of visual representations through the use of minimal synthetic pre-training datasets, we believe that the FDSL framework will be key to discovering the essence of pre-training mechanisms in image recognition.}

\na{\section{Scaling Backwards with a Single Fractal}

This section presents the proposed methodology to explore the minimal requirements for successful pre-training by utilizing purely synthetic images of fractals.
Specifically, we introduce the 1 parameter Fractal as Data (\texttt{1p-frac}), which consists of only a single fractal, with a method to pre-train neural networks on it.
Our key idea is to introduce {\it the locally integrated empirical (LIEP) distribution} $p_{\Delta}$ over perturbed fractal images, enabling pre-training even with a single fractal image. Since the LIEP distribution is designed so that it converges to the empirical distribution $p_{\text{data}}(x) = \delta(x-I)$ of a single image $I$ when the perturbation degree $\Delta \in \mathbb{R}_{\geq 0}$ goes to zero, we can narrow down the support of the distribution by decreasing $\Delta$ as shown in Figure~\ref{fig:shape_aug}\textcolor{red}{a}.
Below, we begin with a preliminary for defining the empirical distributions of two fractal databases, namely FractalDB~\cite{KataokaIJCV2022} and OFDB~\cite{NakamuraICCV2023}.
We then introduce \texttt{1p-frac} with the LIEP distribution.

\subsection{Preliminary}

\noindent\textbf{FractalDB~\cite{KataokaIJCV2022}.} 
Kataoka \textit{et al}. have introduced a method for effectively pre-training neural networks with FractalDB, a set of fractal images generated by the iterated function systems (IFSs).
Specifically, FractalDB $\mathcal{F}$ consists of one million synthesized images: $\mathcal{F} =
\{(\Omega_{c}, \{I_{i}^{c}\}_{i=0}^{M-1})\}_{c=0}^{C-1}$, where $\Omega_{c}$ is an IFS, $I^{c}_{i}$ is a fractal image generated by $\Omega_{c}$, $C = 1{,}000$ is the number of fractal categories, and $M = 1{,}000$ is the number of images per category.
Each IFS characterizes each fractal category $c$ and is defined as follows:
\begin{align}
\Omega_{c} = \{\mathbb{R}^{2}; w_{1}, w_{2}, \ldots, w_{N_{c}}; p_{1}, p_{2}, \ldots, p_{N_{c}} \},
\end{align}
where $w_{j} : \mathbb{R}^{2} \to \mathbb{R}^{2}$ is a 2D affine transformation given by
\begin{align}
w_{j} \left(\bm{v} \right)
=
\begin{bmatrix}
a_j & b_j \\
c_j & d_j
\end{bmatrix}
\bm{v} +
\begin{bmatrix}
e_j\\
f_j
\end{bmatrix}
~(\bm{v} \in \mathbb{R}^{2}),
\end{align}
and $p_{j}$ is a probability mass distribution.
Each fractal image $I^{c}_{i}$ renders a fractal $F = \{\bm{v}_{t}\}_{t=0}^{T} \subset \mathbb{R}^{2}$ into a 2D image, where points $\bm{v}_{t}$ are determined by a recurrence relation that applies affine transformations as $\bm{v}_{t+1} = w_{\sigma_{t}} (\bm{v}_{t})$ for $t = 0, 1, 2, \cdots, T-1$. Here, the initial point is set as $\bm{v}_{0} = (0, 0)^{\top}$, and the index $\sigma_{t}$ is sampled at each $t$ independently from the probability mass distribution $p(\sigma_{t} = j) = p_{j}$.
Pre-training with FractalDB utilizes the cross-entropy loss function:
\begin{align}
\mathcal{L} = -\mathbb{E}_{x,y \sim p_{\text{data}}}[\log p_{\theta}(y|x)],
\end{align}
where $p_{\theta}$ is the category distribution predicted by a neural network and $\theta$ is a set of learnable parameters.
The joint empirical distribution $p_{\text{data}}$ is defined over the dataset as follows:
\begin{align}
p_{\text{data}}(x,y; \mathcal{F}) = \frac{1}{MC} \sum_{i=0}^{M-1} \sum_{c=0}^{C-1} \delta(x-I^{c}_{i}) \delta(y - c)
\end{align}
where $\delta$ is the Dirac's delta function.
Models pre-trained on this dataset perform comparable to those pre-trained on real-world image datasets, such as ImageNet-1k and Places365, in some downstream tasks.

\begin{figure}[t]
\centering
\includegraphics[width=1.0\linewidth]{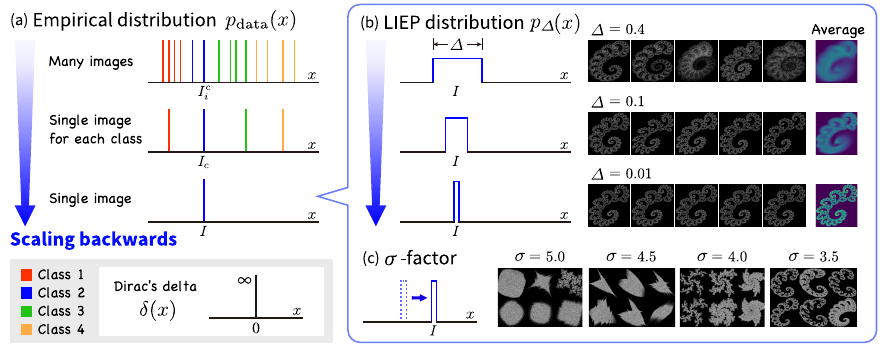}
\caption{
\na{
Scaling backwards from many images to a single synthetic image.
(a) Empirical distribution $p_{\text{data}}$. Colors indicate classes. With a single image, the distribution is given by a single Dirac's delta function.
(b) LIEP distribution $p_{\Delta}$. The support of the distribution narrows as the degree of perturbation $\Delta$ decreases.
(c) $\sigma$-factor for investigating fractal shapes. A small $\sigma$ produces complex fractals.
}
}
\label{fig:shape_aug}
\end{figure}

\noindent \textbf{OFDB~\cite{NakamuraICCV2023}.}
This dataset consists of 1{,}000 fractal images.
Specifically, OFDB $\mathcal{F}_{\text{OF}}$ involves only one representative image per category, {\textit{i.e.},} $\mathcal{F}_{\text{OF}} =
\{\Omega_{c}, I_{c}\}_{c=0}^{C-1}$.
Therefore, the joint empirical distribution reduces to 
\begin{align}
p_{\text{data}}(x,y; \mathcal{F}_{\text{OF}}) = \frac{1}{C} \sum_{c=0}^{C-1} \delta(x-I_{c}) \delta(y - c).
\end{align}
Models pre-trained on this dataset perform comparable to or even better than those pre-trained on FractalDB. This work has shown that there exists a small but essential set of images for visual pre-training.
However, reducing the number of fractals $C$ to less than 1{,}000 degrades the performance.

\subsection{Pre-training with a Single Fractal}

\noindent \textbf{Scaling backwards.}
To further facilitate the analysis of what images are minimally required for successful visual pre-training, we introduce \texttt{1p-frac}, which ultimately reduces the number of IFSs and images to one each as $\mathcal{F}_{\text{OP}} = (\Omega, I)$.
With this dataset, the empirical distribution is given by
\begin{align}
\label{eq:emp_one_image}
p_{\text{data}}(x,y; \mathcal{F}_{\text{OP}}) = \delta(x - I) \delta(y).
\end{align}
However, we notice that with this distribution, training neural networks using cross-entropy loss is not straightforward because $p_{\theta}(y=0|x) \equiv 1~(\forall x)$ gives a trivial solution to the loss minimization problem.
To address this, we introduce
locally perturbed cross entropy (LPCE) loss $\mathcal{L}_{\Delta}$, a variant of the cross entropy loss defined with the LIEP distribution.

\noindent \textbf{\textit{Definition 1.}}
\textit{
Let $I_{\bm{\epsilon}} \in \mathcal{X}$ be a perturbed image, where $\mathcal{X}$ is a set of images, $\bm{\epsilon} \in \mathbb{R}^{d}$ is a small perturbation with $d \in \mathbb{N}_{>0}$ and $I_{\bm{0}} = I$ is the original image.
We define the LIEP distribution by
\begin{align}
\label{eq:distribution}
p_{\Delta}(x, y)
&=
\frac{1}{|\mathcal{R}_{\Delta}|}
\int_{\mathcal{R}_{\Delta}}
\delta(x - I_{\bm{\epsilon}}) \delta(y - \bm{\epsilon}) d \bm{\epsilon}
\end{align}
where $\mathcal{R}_{\Delta} \subset \mathbb{R}^{d}$ is a compact set containing the origin and $|\mathcal{R}_{\Delta}|$ is its volume with order $O(|\Delta|^{d})$.
}

\noindent \textbf{\textit{Definition 2.}}
\textit{
We define the LPCE loss by
\begin{align}
\mathcal{L}_{\Delta} &= -\mathbb{E}_{x, y \sim p_{\Delta}} \left[ \log p_{\theta}(y | x) \right],
\end{align}
where $p_{\Delta}$ is the LIEP distribution.
}

\noindent If $\mathcal{R}_{\Delta}$ is a small hypercube or hypersphere, it is straightforward to see that $p_{\Delta}$ approaches to the empirical distribution of Eq.~(\ref{eq:emp_one_image}) when $\Delta$ goes to zero.
Therefore, this loss allows us to analyze visual pre-training effects by narrowing the support of the distribution around a single image.

With \texttt{1p-frac}, we apply perturbation to the affine transformations. As such, the perturbation $\bm{\epsilon}$ is in \rn{$\mathbb{R}^{6*j}$}.} A perturbed image $I_{\bm{\epsilon}}$ is obtained by the noisy affine transformations:
\begin{align}
w_{j}
\left(\bm{v}; \bm{\epsilon}_j
\right)
=
\left( \begin{bmatrix}
a_{j} & b_{j} & e_{j}\\
c_{j} & d_{j} & f_{j}
\end{bmatrix}
+
\rn{\bm{\epsilon}_j}
\right)
\begin{bmatrix}
\bm{v}\\
1
\end{bmatrix}
\end{align}
where \rn{$\bm{\epsilon}_j \in \mathcal{R}_{\Delta} = [-\Delta/2, \Delta/2]^{6*j} \subset \mathbb{R}^{6*j}$ }is a hypercube with a side length $\Delta$, and
\rn{$|\mathcal{R}_{\Delta}| = \Delta^{6*j}$.}
Note that the numerical integration of Eq.~(\ref{eq:distribution}) is used in practice, which is an approximation obtained by uniformly sampling $L$ points in $\mathcal{R}_{\Delta}$, where we set $L = 1{,}000$ by default.

\noindent \textbf{Visualization.}
Figure~\ref{fig:shape_aug}\textcolor{red}{b} shows examples of perturbed images to compute the LPCE loss. While most of shape differences are indistinguishable to humans, a neural network learns to distinguish the perturbations applied to the single image by minimizing the LPCE loss.

\noindent \textbf{Complexity of $\Omega$.}
We use the $\sigma$-factor proposed by Anderson \textit{et al.}~\cite{AndersonWACV2022} to evaluate the complexity of IFSs.
As shown in Figure~\ref{fig:shape_aug}\textcolor{red}{c},
small values of $\sigma$ produce complex fractal shapes.

\section{Experiments}
\noindent\textbf{Experimental setup.} What is minimal visual \hk{pre-training}, and what properties of images lend themselves to reducing their count? \ryu{In order to} clarify these questions, we conduct experiments following these steps:
\begin{itemize}
    \item \textbf{Exploration study (Tables \ref{tab:fdb-ofdb-opfdb}-\ref{tab:gauss_uniform_dist}):} We verify whether a single \na{fractal is} sufficient for pre-training. \ryu{Moreover}, we compare the extent to which the shapes used in pre-training can be simplified, using object contours, Gaussian, and uniform distributions.
    \item \textbf{Hyperparameter study (Tables \ref{tab:ifs_cataug}-\ref{tab:numof_categories}):} We investigate the effects of three hyperparameters $\Delta$, $\sigma$ and $L$.
    \item \textbf{Scaling study (Table \ref{tab:imagenet-1k-finetuning}):} We compare \texttt{1p-frac} with other large-scale datasets in terms of fine-tuning accuracy on ImageNet-1k.
    \item \textbf{Analysis and discussion (Tables \ref{tab:dataaug}-\ref{tab:affine-elastic-polynomial}):} We discuss data augmentation, computational cost for synthesizing images, and pre-training using a single real-world image.
    \item \textbf{Applications (Tables \ref{tab:main_comparison} and \ref{tab:VTAB}):} We conduct comparisons in various fine-tuning datasets.
\end{itemize}
\noindent\textbf{Implementation details.} \ryu{In order to} verify the effects of pre-training, we measure the accuracy of downstream tasks through fine-tuning. \hk{We use ViT~\cite{DosovitskiyICLR2021} for all experiments. Specifically, we employ ViT-Tiny (ViT-T) in the the exploration and hyperparameter studies, and employ ViT-Base (ViT-B) for the scaling study.} The pre-training parameters and data augmentation methods for pre-training and fine-tuning follow conventional methods in OFDB. In OFDB, data augmentation is based on DeiT. For exploration studies, CIFAR-100 (C100)~\cite{Krizhevsky2009_cifar} and/or ImageNet-100 (IN100)~\cite{RussakovskyIJCV2015} are used as fine-tuning datasets. In the comparisons, we assign seven representative datasets used in OFDB paper, namely C10/C100, Cars~\cite{Krause3DRR2013_cars}, Flowers~\cite{Nilsback08_flowers}, Pascal VOC 2012 (VOC12)~\cite{EveringhamIJCV2015_voc}, Places-30 (P30)~\cite{ZhouTPAMI2017_Places}, IN100. \hk{See \SM{supplementary material} for more detailed experimental settings.}

\subsection{Exploration Study}

\noindent\textbf{Pre-training with \na{\texttt{1p-frac}}.}
\na{Table~\ref{tab:fdb-ofdb-opfdb} compares \texttt{1p-frac} with FracdalDB (1M images, 1k categories) and OFDB (1k images, 1k categories).}
\na{As can be seen}, the pre-training effect \na{of \texttt{1p-frac} is} comparable to \na{that of} FractalDB and OFDB. The number of fractal categories is significantly reduced from $C=1{,}000$ to only $1$.
This shows the effectiveness and \rn{efficiency} of \texttt{1p-frac} and the LPCE loss.

\noindent\textbf{Comparison with \na{SOTA} FDSL datasets.}
\na{Table~\ref{tab:other_fdsl} compares
\texttt{1p-frac} with two SOTA FDSL datasets:
(1) Radial Contour Database (RCDB)~\cite{KataokaCVPR2022}, which consists of 1 million images of synthetic polygons, and (2) Visual Atoms (VA)~\cite{TakashimaCVPR2023}, which consists of 1 million images of parameterized wave functions.
The ``1K'' counterparts in the table reduce the number of images per category to one.
We also applied the LPCE loss to these two datasets by applying perturbations to the radios of a polygon or wave.
As can be seen, \texttt{1p-frac} performs the best when a single image is used for pre-training and outperforms the 1K FDSL datasets.
}

\begin{table*}[t]
\begin{tabular}{cc}
    \begin{minipage}{0.33\hsize}
    \centering
    \caption{\na{Comparison of \texttt{1p-frac} with FractalDB~\cite{KataokaIJCV2022} and OFDB~\cite{NakamuraICCV2023}.
    $^\diamond$ indicates the use of LPCE loss.
    We report top-1 accuracies (\%). 
    }
}
    \vspace{-10pt}
    \label{tab:function}
    \begin{tabular}{l|c|ccc}
        \toprule[0.8pt]
        Method & \#Img & C100 & IN100 \\
        \midrule[0.5pt]
        Scratch & - & 64.2 & 74.9  \\
        FractalDB  & 1M & 81.6 & 88.5 \\
        OFDB &  1k & 84.0 & 88.6  \\
        \rowcolor[gray]{0.8} \texttt{1p-frac}$^\diamond$  & \textbf{1} & \textbf{84.2} & \textbf{89.0} \\
        \bottomrule
    \end{tabular}
    \label{tab:fdb-ofdb-opfdb}
    \end{minipage}

    \hspace{4pt}

	\begin{minipage}{0.33\hsize}
    \centering
    \caption{
\na{Comparison with SOTA FDSL datasets. $^\diamond$ indicates the use of LPCE loss.}
}
    \vspace{-9pt}
    \begin{tabular}{l|c|c|cc}
        \toprule[0.8pt]
        Method  & \#Img & C100  & IN100 \\
        \midrule[0.5pt]
        RCDB & 1M & 81.6 & 88.5  \\
        RCDB & 1K & 80.4 & 87.5  \\
        \rowcolor[gray]{0.925} RCDB$^\diamond$ & \textbf{1} & 82.5 & 87.9 \\
        \midrule[0.5pt]
        \rowcolor[gray]{1.0}
        VA & 1M & 84.9 & 90.3 \\
        VA & 1K & 82.1 & 88.5 \\
        \rowcolor[gray]{0.925}
        VA$^\diamond$ & \textbf{1} & 82.6 & 88.0  \\
        \midrule[0.5pt]
        \rowcolor[gray]{0.8} \texttt{1p-frac}$^\diamond$ & \textbf{1} & 84.2 & 89.0  \\
        \bottomrule
    \end{tabular}
    \label{tab:other_fdsl} 
    \end{minipage}

    \hspace{4pt}

    \begin{minipage}{0.33\hsize}
    \centering
\caption{
Comparison with pre-training with a single noise image.
    }
    \vspace{-10pt}
    \begin{tabular}{lcc}
        \toprule[0.8pt]
        Method & C100  & IN100 \\
        \midrule[0.5pt]
        Gaussian$^\diamond$ & 1.1 & 5.7  \\
        Uniform$^\diamond$ & 2.0 & 71.1 \\
        \rowcolor[gray]{0.8} \texttt{1p-frac}$^\diamond$ & \textbf{84.2} & \textbf{89.0} \\
        \midrule
        \begin{minipage}{20mm}
        \centering
        \includegraphics[width=3.8cm]{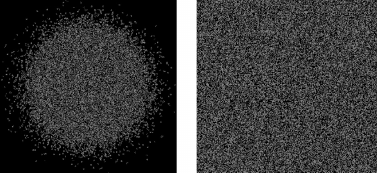}
        \end{minipage} \\

        \bottomrule
    \end{tabular}
    \label{tab:gauss_uniform_dist}
    \end{minipage}
    
\end{tabular}
\end{table*}

\noindent\textbf{Pre-training with \na{a noisy image}.}
\na{
To validate the necessity of a complex shape when pre-training with a single image, we applied the LPCE loss to two noisy images: an image of Gaussian noise and an image of uniform noise. 
\rn{
Since these two noise images are generated by determining the parameters of the parametric distributions, Gaussian and uniform, the LPCE loss can be computed by giving small perturbations to the parameters.}
Table~\ref{tab:gauss_uniform_dist} shows these images and fine-tuning results.
For example, here the $\epsilon$ is added to the mean vector and covariance matrix of the Gaussian distribution. }
The results revealed that \na{noise images} fail in visual pre-training, \na{performing worse than training from scratch.}
This suggests that acquiring fundamental visual representations through pre-training requires an image of \na{a certain structured object}, such as a fractal.


\subsection{\na{Hyperparameter Study}}
\noindent\na{\textbf{Perturbation degree.} Table~\ref{tab:ifs_cataug} shows the results obtained by different perturbation degrees $\Delta$ for the LIEP distribution.}
The results indicate that setting the value to 0.1 yields the highest \na{performance.
Interestingly, the performance significantly improved when $\Delta$ was increased from 0.01 to 0.05.
This suggests that the support of the empirical distribution must have a certain size to obtain positive pre-training effects.
}

\noindent\textbf{$\sigma$-factor.}
\na{Table~\ref{tab:sigma_factor} investigates the effects of the} $\sigma$-factor, \na{which measures the complexity of fractal shapes.
As can be seen, the most complex shape yields the highest performance.
It is worth noting that, even with a fractal of $\sigma = 6.0$, which looks like Gaussian noise, we observed positive pre-training effects.
This suggests that the shapes obtained from IFS are crucial for pre-training.}

\begin{table*}[t]
\centering
\begin{tabular}{ccc}
\begin{minipage}{0.32\hsize}
\caption{\na{Effects of perturbation degree $\Delta$ ($\sigma = 3.5$).}}
\vspace{-6pt}
\centering
\renewcommand{\arraystretch}{0.95}
\setlength{\tabcolsep}{7pt}
\begin{tabular}{c|cc}
\toprule[0.8pt]
$\Delta$ & C100 & IN100 \\
\midrule[0.5pt]
0.001 & 1.2 & 1.9 \\
0.01 & 19.9 & 61.8 \\
0.05 & 83.0 & 88.2 \\
0.1 & \textbf{84.2} & \textbf{89.0} \\
0.2 & 83.4 & 88.5 \\
1.0 & 82.6& 88.1 \\
\bottomrule[0.8pt]
\end{tabular}
\label{tab:ifs_cataug}   
\end{minipage}%
\hfill%
\qquad
\begin{minipage}{0.28\hsize}
\caption{\na{Effects of $\sigma$. \mbox{($\Delta = 0.1$).}}}
	\centering
     \setlength{\tabcolsep}{6pt}
    \begin{tabular}{c|cc}
    \toprule[0.8pt]
    $\sigma$ & C100 & IN100 \\
    \midrule[0.5pt]
    6.0 & 81.3 & \ryu{86.3} \\
    5.0 & 82.2 & 87.1 \\
    4.5 & 81.9 & 86.9 \\
    3.5 & \textbf{84.2} & \textbf{89.0} \\
    4.0 & 82.8 & 87.9 \\
    \bottomrule[0.8pt]
    \end{tabular}
    \label{tab:sigma_factor}
    \end{minipage}%
\hfill%
\qquad
	\begin{minipage}{0.3\hsize}
    \centering
    \caption{\na{Effects of number of sampling points $L$ for numerical integration.}}
    \setlength{\tabcolsep}{6pt}
    \begin{tabular}{l|cc}
        \toprule[0.8pt]
        $L$ & C100  & IN100 \\
        \midrule[0.5pt]
        16  & \ryu{78.7} & \ryu{85.3}  \\
        64  & \ryu{82.3} & \ryu{88.0} \\
        512  & 82.4 & \ryu{88.0} \\
        1000  & \textbf{84.2} & \textbf{\ryu{89.0}} \\
        \bottomrule
    \end{tabular}
    \label{tab:numof_categories}
    \end{minipage}
	\end{tabular}
\end{table*}

\begin{table*}[t]
\begin{tabular}{cc}
	\begin{minipage}{0.60\hsize}
    \centering
    \caption{\hk{Comparison of pre-training datasets on ImageNet-1k fine-tuning. Fine-tuning accuracies calculated by using ViT-Base (B) are listed in the table. 21k indicates the number of categories in each pre-training dataset.}}
    \vspace{-8pt}
    \setlength{\tabcolsep}{8pt}
    \begin{tabular}{lccc} \toprule[0.8pt]
        Pre-training & \#Img & Type & ViT-B\\
        \midrule[0.5pt]
        Scratch & -- & -- & 79.8 \\
        ImageNet-21k & 14M & SL & 81.8 \\
        FractalDB-21k & 21M & FDSL & 81.8 \\
        ExFractalDB-21k & 21M & FDSL & 82.7 \\
        RCDB-21k & 21M & FDSL & 82.4 \\
        VA-21k & 21M & FDSL & 82.7 \\
        OFDB-21k & 21k & FDSL & 82.2 \\
        3D-OFDB-21k & 21k & FDSL & 82.7 \\
        \rowcolor[gray]{0.8} \texttt{1p-frac} (ours) & \textbf{1} & FDSL & 82.1 \\
        \bottomrule[0.8pt]
    \end{tabular}
    \label{tab:imagenet-1k-finetuning}
    \end{minipage}

    \hspace{4pt}
    
    \begin{minipage}{0.35\hsize}
    \centering
    \caption{\hk{Ablation study on data augmentation methods. The DeiT augmentation~\cite{TouvronICML2021} is used as a baseline.}}
    \vspace{-10pt}
    \setlength{\tabcolsep}{6pt}
    \begin{tabular}{l|c} 
    \toprule[0.8pt]
    Method & C100 \\
    \midrule[0.5pt]
    Baseline & 84.2 \\
    \midrule[0.5pt]
    w/o Random Aug.\cite{rand_aug} & 83.4 \\
    w/o Random Crop & \textbf{80.1} \\
    w/o Rand Aspect & 83.6 \\
    w/o Rand Erasing\cite{randm_era} & 84.0 \\
    w/o Mixup\cite{mixup} & 83.6 \\
    w/o Cutmix\cite{cutmix} & 83.3 \\
    w/o Mixup/Cutmix & \textbf{80.4} \\
    w/o Flipping & 84.3 \\
    w/o Color Jittering & 84.2 \\
     \toprule[0.8pt]
    \end{tabular}
    \label{tab:dataaug}
    \end{minipage}
\end{tabular}
\vspace{-10pt}
\end{table*}

\noindent\textbf{\na{Numerical integration.}} 
Tables~\ref{tab:numof_categories} reduces the number of sampling points $L$ of the numerical integration when approximately computing loss with Eq.~(\ref{eq:distribution}).
We see that larger numbers result in high performance.
Interestingly, even with $L=16$, the pre-training effect was positive (better than training from scratch).

\subsection{\na{Scaling Study}}

\noindent\textbf{\hk{ImageNet-1k fine-tuning (Table~\ref{tab:imagenet-1k-finetuning}}).} 
While this work discusses minimal requirements for visual pre-training, increasing the approximation precision of the numerical integration could further improve the performance and benefit training of large models.
In Table~\ref{tab:imagenet-1k-finetuning}, we applied the LPCE loss with $L=21{,}000$ to the ViT-B model and compare fine-tuning accuracy on ImageNet-1k.
Surprisingly, and despite using a single fractal as data, our ViT-B pre-trained on \texttt{1p-frac} outperforms pre-training on ImageNet-21k.
This shows the strength of our approach in particular because we simply keep the same fine-tuning protocols as used for the original Imagenet-21k to ImageNet-1k transfer proposed in~\cite{DosovitskiyICLR2021}, putting us at a potential disadvantage.


\subsection{\na{Analysis and Discussion}}
\noindent\textbf{Relationship with data augmentation (Table~\ref{tab:dataaug}).} We consider that the variation in image patterns also depends on data augmentation. While traditional methods primarily rely on the data augmentation techniques described in the DeiT paper~\cite{TouvronICML2021}, we investigated which data augmentation methods contribute significantly by excluding one or two techniques from the base data augmentation methods and observing the changes in fine-tuning accuracy \hk{on C100 dataset}. The experimental results in Table~\ref{tab:dataaug} reveal that excluding Random Cropping (RandomCrop) significantly deteriorates accuracy, 
\hk{suggesting that omitting parts of the image region and causing partial loss facilitates the acquisition of the ability to focus on any part without relying on grasping the overall shape,}
greatly reducing the pre-training effect. Furthermore, when both Mixup and Cutmix are excluded, a degradation in accuracy is observed, indicating that mixing images and categories enables better feature recognition.

\begin{table*}[t]
\begin{tabular}{cc}
	\begin{minipage}{0.40\hsize}
    \centering
\caption{\na{Dataset construction time (hours).
FractalDB (1M images), OFDB (1k images) and \texttt{1p-frac} (1k perturbations for one image) are compared.} \hk{The processing time is shown separately for fractal category search (Search), image rendering (Render), and total time (Total).}}
\vspace{-5pt}
\scalebox{1.0}{
    \begin{tabular}{lccc} \toprule[0.8pt]
    Dataset & Search & Render & Total \\
    \midrule[0.5pt]
    FractalDB & 2.37 & 16.86 & 19.23 \\
    OFDB & 2.37 & 0.41 & 2.78 \\
    \texttt{1p-frac} & 0.0022 & 0.036 & 0.0382 \\
    \toprule[0.8pt]
    \end{tabular}
    }
    \label{tab:processing_time}
    \end{minipage}

    \hspace{4pt}
    
    \begin{minipage}{0.55\hsize}
    \centering 
    \includegraphics[width=0.95\linewidth]{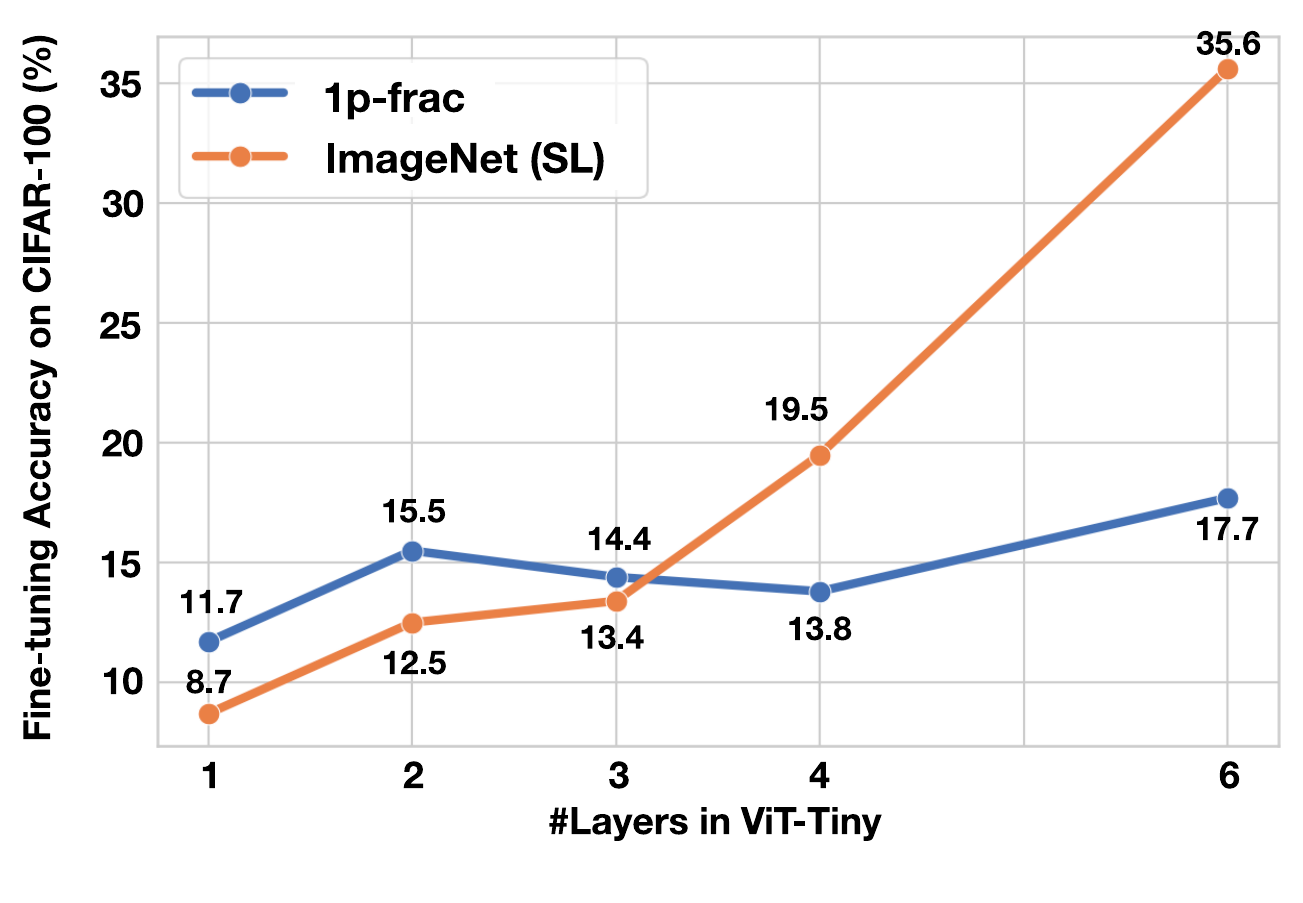}
    \captionof{figure}{\hk{Linear probing following the experiments from \cite{asano2020a}. We verified the performance with \{1, 2, 3, 4, 6\} layers in the ViT-Tiny model on C100 dataset. Here, 2 and 4 show the use of up to 1--2 and 1--4 layers.}}
    \label{fig:layerwiselinearprobing}
    \end{minipage}
\end{tabular}
\vspace{-15pt}
\end{table*}

\hk{
\noindent\textbf{Dataset construction times (Table~\ref{tab:processing_time}).} 
We calculated the processing times in fractal category search and image rendering. By comparing to the conventional approaches, our \texttt{1p-frac} recorded much faster category search (0.0022 hours nearly equals to 8 seconds) and image rendering (0.036 hours nearly equals to 129 seconds). \texttt{1p-frac} requires only one category with a parameter set. This very simple procedure leads to efficiently construct a minimum requirement for pre-training in image recognition.
}

\hk{
\noindent\textbf{Linear probing (Figure~\ref{fig:layerwiselinearprobing}).} 
To compare supervised ImageNet-1k pre-training with our \texttt{1p-frac} pre-training in the context of early layer visual representations, we executed a linear probing experiment on the C100 dataset. In the ViT-T model with \{1, 2, 3\} layers, the \texttt{1p-frac} pre-trained model outperformed the model pre-trained on ImageNet-1k. This demonstrates the superior quality of early, low-level representations learned from our dataset compared to ImageNet.
In effect, layers up to layer 3 in ViT-T can be sufficiently pre-trained and frozen using our \texttt{1p-frac}.
}

\begin{figure}[t]
    \centering 
    \includegraphics[width=1.0\linewidth]{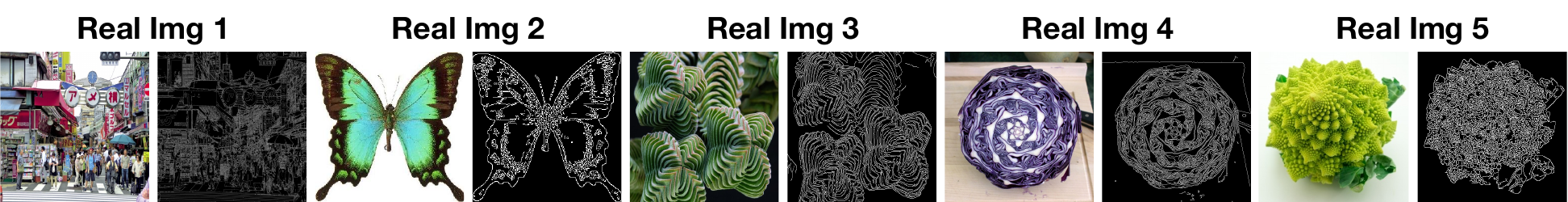}
    \vspace{-15pt}
    \caption{\hk{The five different images used in the pre-training with a single real image and shape augmentation.}}
    \label{fig:real-image-transformation}
    \vspace{-5pt}
\end{figure}

\noindent\textbf{\hk{Pre-training with a single real image and shape augmentation}.}
\begin{wrapfigure}[8]{r}{0.6\textwidth} 
  \centering
  \vspace{-0.9cm}
  \includegraphics[width=0.6\textwidth]{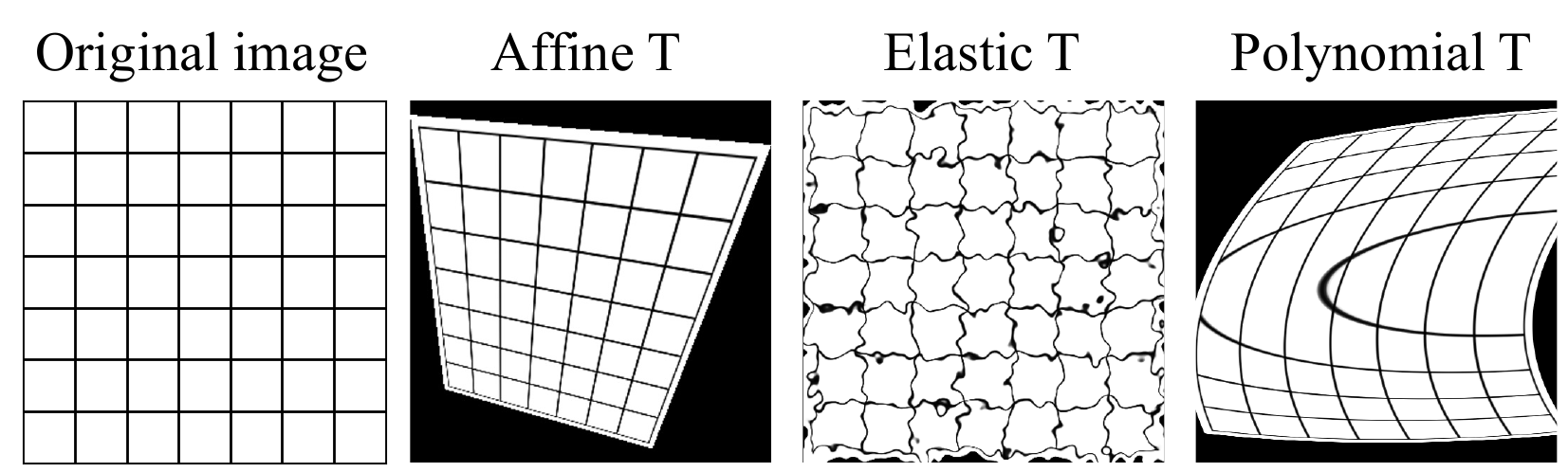}
  \caption{Shape augmentation with \hk{three geometry transformations} from a single real image.}
  \label{fig:affine-elastic-polynomial-transformation}
  \vspace{0cm}
\end{wrapfigure}
\hk{
It was found that pre-training effects comparable to a dataset of one million real-world images can be achieved with \texttt{1p-frac}, but is it possible to perform pre-training similarly with a real image? 
In this section, we verify whether visual features can be acquired through pre-training using a single real image (each image in Figure~\ref{fig:real-image-transformation}) with the LPCE loss using image representations similar to FDSL (Canny~\cite{canny} images) and geometric transformations for perturbation (\{affine, elastic, polynomial\} transformations in Figure~\ref{fig:affine-elastic-polynomial-transformation}). 
The results of pre-training effects with $L=1,000$
are shown in Tables~\ref{tab:rgb-vs-canny} and \ref{tab:affine-elastic-polynomial}. Note that the results in these tables represent the average score on C100 over the five real images in Figure~\ref{fig:real-image-transformation}. See the supplementary material for results for each of image.
According to the results, we find that contour-emphasized Canny images omitting color information and affine transformation perturbation yield better pre-training, as measured by the fine-tuning accuracies. The operations are similar to the classification of labeled fractal images via IFS.
%
}

\begin{table*}[t]
    \begin{minipage}{0.48\hsize}
    \centering
    \caption{
    Comparison of pre-training effect between RGB and Canny images when shape augmentation is applied. The shape augmentation uses affine transformations.
    }
    \setlength{\tabcolsep}{8pt}
    \begin{tabular}{cc}
    \toprule
    Image & Mean Accuracy \\
    \midrule[0.5pt]
    RGB image & 81.6\\
    Canny image & \textbf{82.2}\\
    \bottomrule
    \end{tabular}
    \label{tab:rgb-vs-canny}
    \end{minipage}
\hfill
	\begin{minipage}{0.48\hsize}
    \centering
    \caption{
    Comparison of pre-training effects of affine, elastic, and polynomial image transformations.
    Real Img type with Canny applied to the images.
    }
    \setlength{\tabcolsep}{6pt}
    \vspace{-3pt}
    \begin{tabular}{cc}
    \toprule
    Transformation & Mean Accuracy \\
    \midrule[0.5pt]
    Affine trans. & \ryu{\textbf{81.9}} \\
    Elastic trans. & \ryu{37.0} \\
    Polynomial trans. & \ryu{81.3} \\
    \bottomrule
    \end{tabular}
    \label{tab:affine-elastic-polynomial}
    \end{minipage}

\vspace{-10pt}
\end{table*}

\begin{table*}[t]
    \centering
    \caption{\hk{Comparison among pre-training methods in fine-tuning accuracies. Best values at each dataset scale are in bold. ViT-T is used for all experiments. \#OrgImg indicates number of collected images in real-image datasets or number of original images used in synthetic datasets. Type shows supervised types within supervised learning using cross-entropy loss (SL), self-supervised learning using DINO~\cite{CaronICCV2021_dino} or MAE~\cite{HeCVPR2022} (SSL:D/SSL:M), and formula-supervised learning (FDSL).}}
    \vspace{-5pt}
    \scalebox{0.89}{
    \begin{tabular}{lccccccccc|c} \toprule[0.8pt]
        Pre-training & \#OrgImg & Type & C10 & C100 & Cars & Flowers & VOC12 & P30 & IN100 & Mean \\\midrule[0.5pt]
        Scratch & -- & -- &  80.3 & 62.1 & 13.7 & 71.4 & 56.2 & 76.2 & 74.8 & 62.1 \\
        \midrule[0.5pt]
        Places-365~\cite{ZhouTPAMI2017_Places} & 1.80M & SL & 97.6 & 83.9 & 89.2 & 99.3 & 84.6 & -- & 89.4 & -- \\
        ImageNet-1k~\cite{DengCVPR2009_ImageNet} & 1.28M & SL & \textbf{98.0} & 85.5 & \textbf{89.9} & \textbf{99.4} & \textbf{88.7} & 80.0 & -- & --  \\
        ImageNet-1k~\cite{DengCVPR2009_ImageNet} & 1.28M & SSL:D & 97.7 & 82.4 & 88.0 & 98.5 & 74.7 & 78.4 & 89.0 & 86.9 \\
        ImageNet-1k~\cite{DengCVPR2009_ImageNet} & 1.28M & SSL:M & 97.4 & \textbf{85.8} & 86.6 & 96.3 & 83.3 & \textbf{80.2} & \textbf{90.0} & \textbf{88.5} \\
        PASS~\cite{AsanoNeurIPS2021_pass}& 1.43M & SSL:D & 97.5 & 84.0 & 86.4 & 98.6 & 82.9 & 79.0 & 82.9 & 87.8 \\
        FractalDB-1k~\cite{KataokaIJCV2022} & 1.00M & FDSL & 96.8 & 81.6 & 86.0 & 98.3 & 80.6 & 78.4 & 88.3 & 87.1 \\
        RCDB-1k~\cite{KataokaCVPR2022} & 1.00M & FDSL & 97.0 & 82.2 & 86.5 & 98.9 & 80.9 & 79.7 & 88.5 & 87.6 \\
        \midrule[0.5pt]
        ImageNet-1k~\cite{NakamuraICCV2023} & $1{,}000$ & SL & 94.3 & 76.9 & 57.3 & 94.8 & 73.8 & 78.2 & 84.3 & 79.9 \\
        ImageNet-1k~\cite{NakamuraICCV2023} & $1{,}000$ & SSL:D & 94.9 & 78.0 & 71.2 & 94.6 & 75.5 & 78.6 & 84.9 & 82.5 \\
        OFDB-1k~\cite{NakamuraICCV2023}& $1{,}000$ & FDSL & 96.9 & 84.0 & 84.5 & 97.1 & 79.9 & 79.9 & 88.0 & 87.2 \\
        3D-OFDB-1k~\cite{NakamuraICCV2023} & $1{,}000$ & FDSL & 97.1 & 83.8 & 85.5 & \textbf{98.4} & 80.8 & 80.0 & 89.1 & 87.8 \\
        OFDB-1k w/ Aug.~\cite{NakamuraICCV2023} & $1{,}000$ & FDSL & \textbf{97.2} & \textbf{85.3} & \textbf{87.6} & 98.3 & \textbf{81.4} & \textbf{80.4} & \textbf{89.5} & \textbf{88.5} \\
        3D-OFDB-1k w/ Aug.~\cite{NakamuraICCV2023} & $1{,}000$ & FDSL & 97.0 & 84.7 & 85.6 & 98.3 & 81.2 & 79.8 & 88.9 & 87.9 \\
        \midrule[0.5pt]
        1-image SSL~\cite{asano2020a} & \textbf{1} & SSL:D & 95.7 & 79.5 & 73.1 & 93.8 & 69.0 & 80.4 & 88.7 & 82.8 \\
        1-image SSL~\cite{asano2020a} & \textbf{1} & \ryu{SSL:M} & \ryu{\textbf{97.2}} & \ryu{\textbf{84.7}} & \ryu{82.1} & \ryu{94.6} & \ryu{78.5} & \ryu{80.3} & \ryu{\textbf{89.0}} & \ryu{86.6} \\
        \rowcolor[gray]{0.8} \texttt{1p-frac} (ours) & \textbf{1} & FDSL & 96.9 & 84.2 & 84.5 & 97.4 & 80.5 & \textbf{80.6} & \textbf{89.0} & 87.5 \\
        \rowcolor[gray]{0.8} \texttt{1p-frac} w/ Aug. (ours) & \textbf{1} & FDSL & 96.5 & \textbf{84.7} & \textbf{87.0} & \textbf{98.1} & \textbf{80.9} & 80.5 & 88.9 & \textbf{88.2} \\
        \bottomrule[0.8pt]
    \end{tabular}
    }
    \vspace{-10pt}
    \label{tab:main_comparison}
\end{table*}

\begin{table*}[t]
\centering
\caption{\hk{Performance comparisons on five fine-tuning datasets: Retinopathy (Retino)~\cite{RetinopathyKaggle2015}, Resisc45~\cite{ChengIEEE2017_resisc45}, Camelyon~\cite{VeelingMICCAI2018}, CLEVR-Count (CLEVR-C)~\cite{JohnsonCVPR2017_CLEVR}, and sNORB-Azim (sNORB-A)~\cite{LeCunCVPR2004}. Best and second-best values are in bold and underlined, respectively.}}
\begin{tabular}{llcc|ccc|cc}
\toprule[0.8pt]
                             &              & \#OrgImg & Type & Retino & Resisc45 & Camelyon & CLEVR-C & sNORB-A \\ 
\midrule[0.5pt]
 & Scratch & -- & --      &  66.9         & 89.2      &  74.4   & 46.4      & 11.5\\
                             & ImageNet-1k~\cite{DengCVPR2009_ImageNet} & 1.28M & SL &  \textbf{79.1}     & \textbf{97.0}     &  \underline{82.7}   & \underline{89.3}        & \textbf{29.2}  \\
                              & ImageNet-1k~\cite{DengCVPR2009_ImageNet}  & 1.28M & SSL:M & 76.7     & 95.6     & \textbf{83.9}   & \textbf{89.5}      & 26.3\\
                             \rowcolor[gray]{0.8}& \texttt{1p-frac} (ours)  & \textbf{1} & FDSL & \underline{77.9}         & \underline{95.8}      &   \underline{82.7}  & 86.2        & \underline{28.9}\\ 
\bottomrule[0.8pt]
\vspace{-15pt}
\label{tab:VTAB}
\end{tabular}
\end{table*}

\hk{
\subsection{Applications}
\noindent\textbf{Fine-tuning on commonly used datasets (Table~\ref{tab:main_comparison}).} We compare fine-tuning accuracies in three types of dataset configurations. For supervised learning (SL) we compare pretraining on ImageNet-1k and Places-365, for self-supervised learning (SSL) we compare  DINO~\cite{CaronICCV2021_dino} and masked auto-encoder (MAE)~\cite{HeCVPR2022} pre-trained on ImageNet and PASS, and models trained on the fractal datasets FDSL on FractalDB-1k and RCDB-1k. Following the setting of one-instance per category~\cite{NakamuraICCV2023}, we also evaluate using random patch augmentation (``w/ Aug.''). 
We compare our \texttt{1p-frac} with 1-image SSL~\cite{asano2020a} with DINO and MAE (we use Real Img \#1 of Figure~\ref{fig:real-image-transformation}). 
The results show that our \texttt{1p-frac} pre-training with one single image demonstrated high fine-tuning accuracies (87.5\% and 88.2\% with and without random patch augmentation) which are close to the performance of self-supervised ImageNet-1k pre-training with MAE that utilizes 1.28M images and yields 88.5\%. Surprisingly, despite using only one fractal image with perturbations, our \texttt{1p-frac} without random patch augmentation has demonstrated fine-tuning accuracies that are equal to or even higher than that of conventional OFDB-1k/FractalDB-1k, which uses $1{,}000$ parameter sets with fractal categories and $1{,}000$/$1{,}000{,}000$ images: \texttt{1p-frac} obtains mean performances of 87.5\%  vs OFDB-1k 87.2\% vs FractalDB-1k 87.1\%.

\noindent\textbf{Fine-tuning on specialized and structured datasets in VTAB~\cite{ZhaiarXiv2019_VTAB} (Table~\ref{tab:VTAB}).} We also verify performance on a couple of specialized and structured datasets in VTAB (visual task adaptation benchmark; Retinopathy~\cite{RetinopathyKaggle2015}, Resisc45~\cite{ChengIEEE2017_resisc45}, Camelyon~\cite{VeelingMICCAI2018}, CLEVR-Count~\cite{JohnsonCVPR2017_CLEVR}, and sNORB-Azim~\cite{LeCunCVPR2004}). The experiment is more suitable to evaluate and compare the pre-training methods since the image domains and their labels are distinct from both real and fractal images on ImageNet and \texttt{1p-frac}. According to Table~\ref{tab:VTAB}, our \texttt{1p-frac} pre-trained model performed relatively similar fine-tuning accuracies on all listed datasets in the table. By comparing to self-supervised ImageNet-1k with MAE, our \texttt{1p-frac} surpassed the fine-tuning accuracies on three (Retinopathy, Resisc45, and sNORB-Azim) out of the five datasets. 
}

\section{Conclusion and Discussion}
This paper examined what a minimal dataset for synthetic pre-training might look like. We proposed the 1-parameter Fractal as Data (\texttt{1p-frac}) dataset that succeeds in pre-training even with a single fractal by utilizing the locally perturbed cross entropy (LPCE) loss.
The following findings were presented:

\noindent\textbf{What is \hk{a minimal synthetic pre-training dataset}?} In this paper, we applied IFS to generate labeled images. In this context, preparing a single set of parameters $a-f$ in IFS is the least informative. It recorded fine-tuning accuracy nearly equivalent to conventional methods such as FractalDB and OFDB \hk{(Tables~\ref{tab:scaling_backwards} and \ref{tab:function})}, succeeding in pre-training with minimal image pattern utilization. Moreover, it is not that any perturbation in image space is acceptable, as Gaussian/uniform distributions failed in pre-training (Table~\ref{tab:gauss_uniform_dist}). It clarified the importance of geometric transformations with certain regularities like fractal geometry or contour shapes (Table~\ref{tab:other_fdsl}).

\noindent\textbf{What properties of images lead to image reduction?} In \texttt{1p-frac}, we investigated shape changes and pre-training effects by adopting the LPCE loss and the $\sigma$-factor that contribute to fractal shape variation. It was revealed that pre-training efficacy varies with shape variation (Tables~\ref{tab:ifs_cataug}, \ref{tab:sigma_factor}). For real images, it is not just about including more edge components, but about including affine transformations, 
which closely resembles the classification of labeled fractal images through IFS (Tables~\ref{tab:rgb-vs-canny} and \ref{tab:affine-elastic-polynomial}).
By extensively ablating data augmentation showed, we demonstrated the essential role of random cropping and MixUp.  

\noindent\textbf{Pre-training effects with the proposed method.} \hk{Supervised pre-training on ImageNet} was more accurate for datasets where object labels are assigned to real images (Table~\ref{tab:main_comparison}), but in VTAB's Specialized/Structured datasets (Table~\ref{tab:VTAB}), performances were \hk{relatively} close. \hk{On the other hand, our \texttt{1p-frac} pre-trained ViT performed a similar mean accuracy by comparing to MAE self-supervised ImageNet pre-training (MAE 88.5 vs. our \texttt{1p-frac} w/ Aug. 88.2).}
Furthermore,
\na{increasing the number of sampling points $L$} showed performance improvement even with 21k categories and demonstrated good performance in ImageNet-1k fine-tuning \hk{(Table~\ref{tab:imagenet-1k-finetuning})}, indicating the potential for building large models with limited data resources.

\noindent\textbf{Broader impact.}
Our proposed dataset \texttt{1p-frac} does not suffer from licensing or ethical issues such as biases and has a large potential to serve as a clean pretraining dataset. 
While not investigated in this paper, this approach also opens the possibility for  significant gains in training speed by keeping the data on the GPU and applying transformations there, removing the GPU-CPU transfer bottleneck.

\noindent\textbf{Limitation.} In this paper, we have explored achieving a minimal synthetic pre-training in image recognition based on fractal geometry with IFS (a single fractal, and their perturbations). 
Despite this, we believe it is also necessary to find a similarly minimal pre-training dataset containing real images -- for some applications and increased interpretability. This might allow for quick and clear calibration of pre-training, similar to how calibration images are used in photography. We did not investigate this direction in this paper but leave this for future work.

\section*{Acknowledgements}
\vspace{-7pt}
Computational resource of AI Bridging Cloud Infrastructure (ABCI) provided by National Institute of Advanced Industrial Science and Technology (AIST) was used.

\vspace{-7pt}

%
%
\bibliographystyle{splncs04}
\bibliography{egbib}

\end{document}